\documentclass[sigconf]{acmart}
\usepackage{booktabs} 
\usepackage{array}

\setcopyright{rightsretained}

\acmDOI{10.475/123_4}

\acmISBN{123-4567-24-567/08/06}

\acmConference[WWW'2018]{}{April 2018}{Lyon, France} 
\acmYear{2018}
\copyrightyear{2018}

\acmArticle{4}
\acmPrice{15.00}

\begin{document}
\renewcommand\arraystretch{1.3}
\title{Netizen-Style Commenting on Fashion Photos: Dataset and Diversity Measures}

 \author{Wen Hua Lin, Kuan-Ting Chen, Hung Yueh Chiang and Winston Hsu}
 \affiliation{%
   \institution{National Taiwan University, Taipei, Taiwan}
   \href{mailto:q868686qq@gmail.com}{q868686qq@gmail.com}, \href{mailto:ktchen@cmlab.csie.ntu.edu.tw}{ktchen@cmlab.csie.ntu.edu.tw}, \href{mailto:kenny5312012@gmail.com}{kenny5312012@gmail.com}, \href{mailto:whsu@ntu.edu.tw}{whsu@ntu.edu.tw}
}







\begin{abstract}
Recently, deep neural network models have achieved promising results in image captioning task. Yet, ``vanilla'' sentences, only describing shallow appearances (e.g., types, colors), generated by current works are not satisfied netizen style resulting in lacking engagements, contexts, and user intentions. To tackle this problem, we propose {\it Netizen Style Commenting (NSC)}, to automatically generate characteristic comments to a user-contributed fashion photo. We are devoted to modulating the comments in a vivid ``netizen'' style which reflects the culture in a designated social community and hopes to facilitate more engagement with users. In this work, we design a novel framework that consists of three major components: (1) We construct a large-scale clothing dataset named NetiLook, which contains 300K posts (photos) with 5M comments to discover netizen-style comments. (2) We propose three unique measures to estimate the diversity of comments. (3) We bring diversity by marrying topic models with neural networks to make up the insufficiency of conventional image captioning works. Experimenting over Flickr30k and our NetiLook datasets, we demonstrate our proposed approaches benefit fashion photo commenting and improve image captioning tasks both in accuracy and diversity.

\end{abstract}

%


\keywords{Fashion; Image Captioning; Commenting; Diversity; Deep Learning; Topic Model}

\maketitle

\begin{figure}[h!]
\centering
\includegraphics[width=84mm]{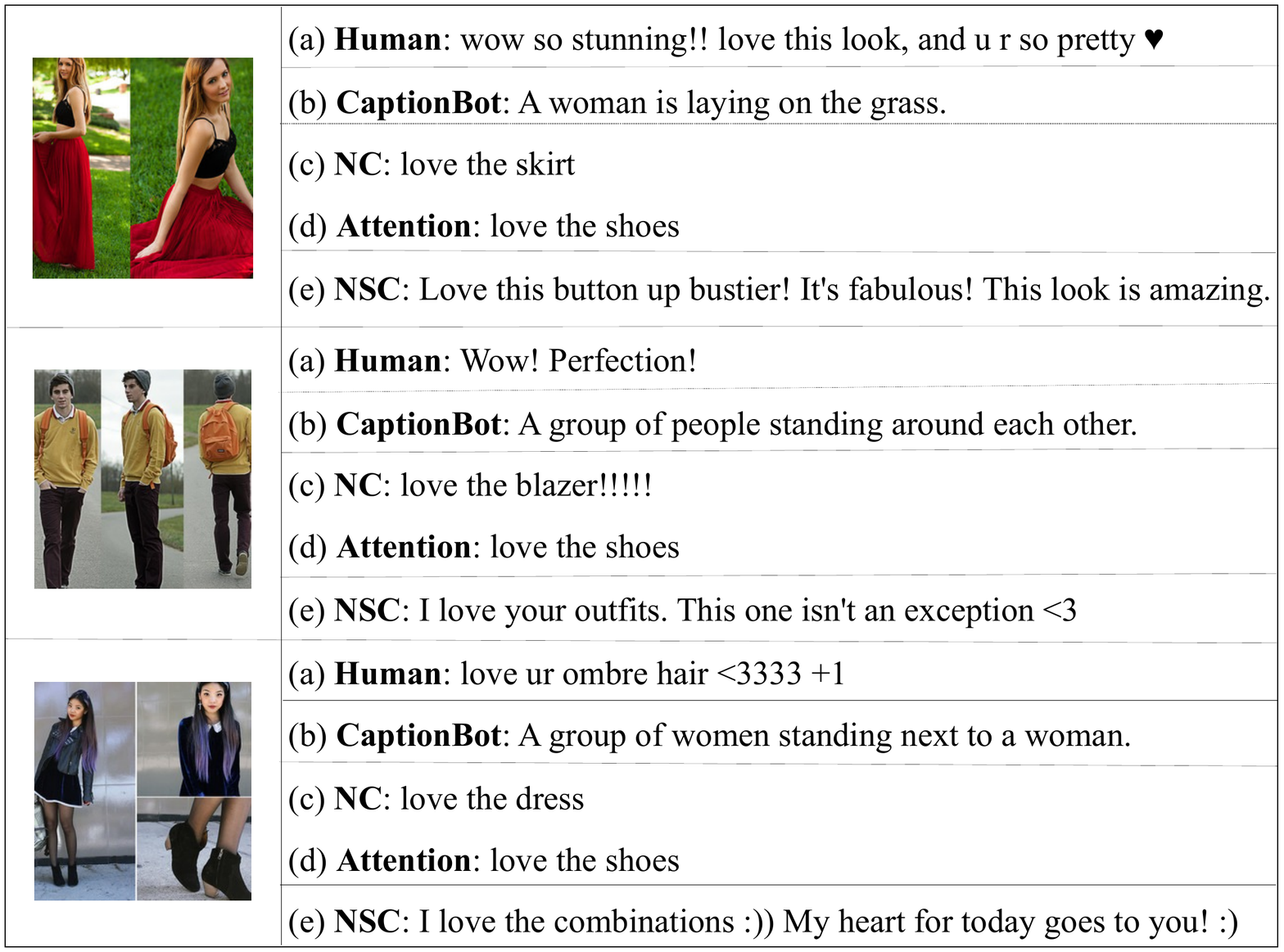}
\caption{Five sentences for each image from distinct commenting (captioning) methods. (a) One of the users' comments (i.e., ground truth) randomly picked from the post (photo) in our collected NetiLook dataset. (b) The sentences from Microsoft CaptionBot. (c) The results from neural image caption generation (NC)~\cite{vinyals2015show} (d) The results from neural image caption generation with visual attention (Attention)~\cite{xu2015show}. (e) Our proposed NSC. It marries style-weight to achieve vivid netizen style results.
}
\label{fig:example1}
\end{figure}

\section{Introduction\label{sec:intro}}
  In accordance with~\cite{simo2015neuroaesthetics}, fashion has a vital impact on our society because clothing typically reflects a person's social status. This is also expected in the growing online retail sales, reaching 529 billion dollars in the US, and 302 billion euros in Europe by 2018 \cite{forbes}. Still today, people either wear up their new clothes or upload their new clothing photo on social media to receive comments of new clothes. However, dressing inappropriately sometimes causes embarrassing. Therefore, people tend to know whether they dress properly beforehand. As the promising results achieved by image captioning, the problem could be solved by fashion image captioning works, which automatically describe the outfit with netizen-like comments.

Whereas, image captioning \cite{donahue2015long}\cite{fang2015captions}\cite{ karpathy2015deep}\cite{mao2014deep}\cite{vinyals2015show}\cite{xu2015show} is still a challenging and under researching topic despite 
deep learning developing rapidly in recent years. To generate a human-like captioning, machines not only recognize objects in an image but express their relationships in natural language, such as English. Large corpora of paired images and descriptions, such as MS COCO~\cite{lin2014microsoft} and Flickr30k~\cite{rashtchian2010collecting} are proposed to address the problem. Several deep recurrent neural network models are devised to follow the datasets and reach promising results. However, modern methods only
focus on optimizing metrics used in machine translation,
which causes absence of diversity --- producing conservative
sentences. These sentences can achieve good scores in machine translation
metrics but are short of humanity. Compared with human comments as shown in Figure~\ref{fig:example1} (a), due to the limitation of training data, current methods (e.g., Figure~\ref{fig:example1} (b)) merely describe ``vanilla'' sentences with low utilities, which are merely describing the shallow and apparent appearances (e.g., color, types) in photos and generate meaningless bot tokens to users --- lacking engagement, contexts, and feedbacks for user intentions, especially in the circumstances of online social media.

  In order to generate human-like online comments (e.g, clothing style) for fashion photos, we collect a large corpus of paired user-contributed fashion photos and comments, called NetiLook, from an online clothing style community. To the best of our knowledge, our collected NetiLook is the largest fashion comment dataset. In our experiment on NetiLook, we found that these methods overfit to a general pattern, which makes captioning results insipid and banal (e.g., ``love the ...'') (cf., Figure~\ref{fig:example1} (c) and (d)). Therefore, to compensate for the deficiency, we propose integrating latent topic models with state-of-the-art methods and make the generated sentences vivacious (cf., Figure~\ref{fig:example1} (e)). Besides, for evaluating diversity, we propose three novel measures to quantize variety.

  For richness and diversity in text content, we propose a novel method to automatically generate characteristic fashion photo comments for user-contributed fashion photos by marrying \textit{style-weight} (cf., Section~\ref{sec:styleCaptioning}) from topic discovery models (i.e., latent Dirichlet allocation (LDA) \cite{blei2003latent}) with neural networks to achieve diverse comments with vivid ``netizen'' style. We look forward the breakthrough will foster further applications in social media, online customer services, e-commerce, chatbot developments, etc. It will be more exciting, in the very near future; for example, if the solution can work as an agent (or expert) in a living room and can comment for a user as testing the outfit in front of the mirror. To sum up, our main contributions are as follows:

\begin{itemize}
\item To our best knowledge, this is the first work to address the diverse measures of photo captioning in a large-scale fashion commenting dataset (cf., Section \ref{sec:intro}-\ref{sec:rw}). 
\item We collect a brand new large-scale clothing dataset, NetiLook, which contains 300K posts (photos) with 5M comments (cf., Section \ref{sec:dataset}).
\item We investigate the diversity of clothing captioning and propose three measures to estimate the diversity (cf., Section \ref{sec:diversityMeasures}).
\item We leverage and investigate the merit of latent topic models, which is able to make up the insufficiency of conventional image captioning works (cf., Section \ref{sec:methods}).
\item We demonstrate that our proposed approach significantly benefits fashion photo commenting and improves image captioning task both in accuracy and diversity over Flickr30k and NetiLook datasets (cf., Section \ref{sec:experiments}).
\end{itemize}

\section{Related Work\label{sec:rw}}
Image captioning which automatically describes the content of an image with properly formed sentences enables many important applications such as helping visually impaired users and human-robot interaction. According to~\cite{anne2016deep}\cite{venugopalan2016captioning}, a CNN-RNN framework, taking high-level features extracted from a deep convolution neural network (CNN) as an input for a recurrent neural network (RNN) to generate a complete sentence in natural language, has performed promisingly in image captioning tasks during the last few years. For example,~\cite{vinyals2015show} is an end-to-end CNN model followed by language generation of RNN. It was able to produce a  grammatically correct sentence in natural language from an input image. 

Following CNN-RNN frameworks, attention-based models (\cite{xu2015show}, \cite{lu2016knowing}) were proposed. As human beings put different attentions at distinct objects while watching a photograph, attention-based models allow machines to put diverse weights on salient features. Compared with taking high-level representations of a whole image as input, attention-based models are able to dynamically weight various parts of images. Especially, when a lot of objects appear in an image, attention-based models can give more thorough captions~\cite{xu2015show}.

Presently, state-of-the-art works are majorly attention-based models (\cite{mun2017text}, \cite{li2017image}) because they focus on correctness of descriptions. \cite{chen2017reference} assigned different weights to different words for fixing misrecognition. \cite{liu2017attention} focused on evaluating the correctness of attention in neural image captioning models. 

While applying current methods to generate comments, the demand for diversity is unveiled. Compared with depicting images, giving comments is more challenging because it needs to not only understand images but take care of engagement with users. To generate vivid comments, diversity is necessary. Besides commenting, diversity is also important in other areas (e.g., information retrieval \cite{chapelle2012large}). In \cite{kannan2016smart}, to increase the utility of automatically generated response options of email, diversity is essential. Moreover, in building general-purpose conversation agents, which are required for intelligent agents' interaction with humans in natural language, diversity is also requisite. Therefore, we blend topic models with conventional methods to complement the diversity part of them. 

Meanwhile, there has been increasing interest in clothing product analysis from the computer vision and multimedia communities. Most existing fashion analysis works focused on the investigation of the clothing attributes, such as clothing parsing (\cite{RW:LiuCVPR12}, \cite{RW:TMM14}, \cite{RW:liu2016deepfashion}), fashion trend (\cite{hidayati:TrendNY}, \cite{Chen:FashionPrada}) and clothing retrieval (\cite{RW:KiapourICCV2015}, \cite{RW:LiuMM12}). In contrast to other works, we develop a novel framework that can leverage the learned Netizen-style embedding for commenting on fashion photos. Moreover, to our best knowledge, this is the first work to address the diverse measures of photo captioning in an all-time large-scale fashion commenting dataset. We detail the dataset and our method in the following sections.

\begin{figure}[t]
\centering
\includegraphics[width=86mm]{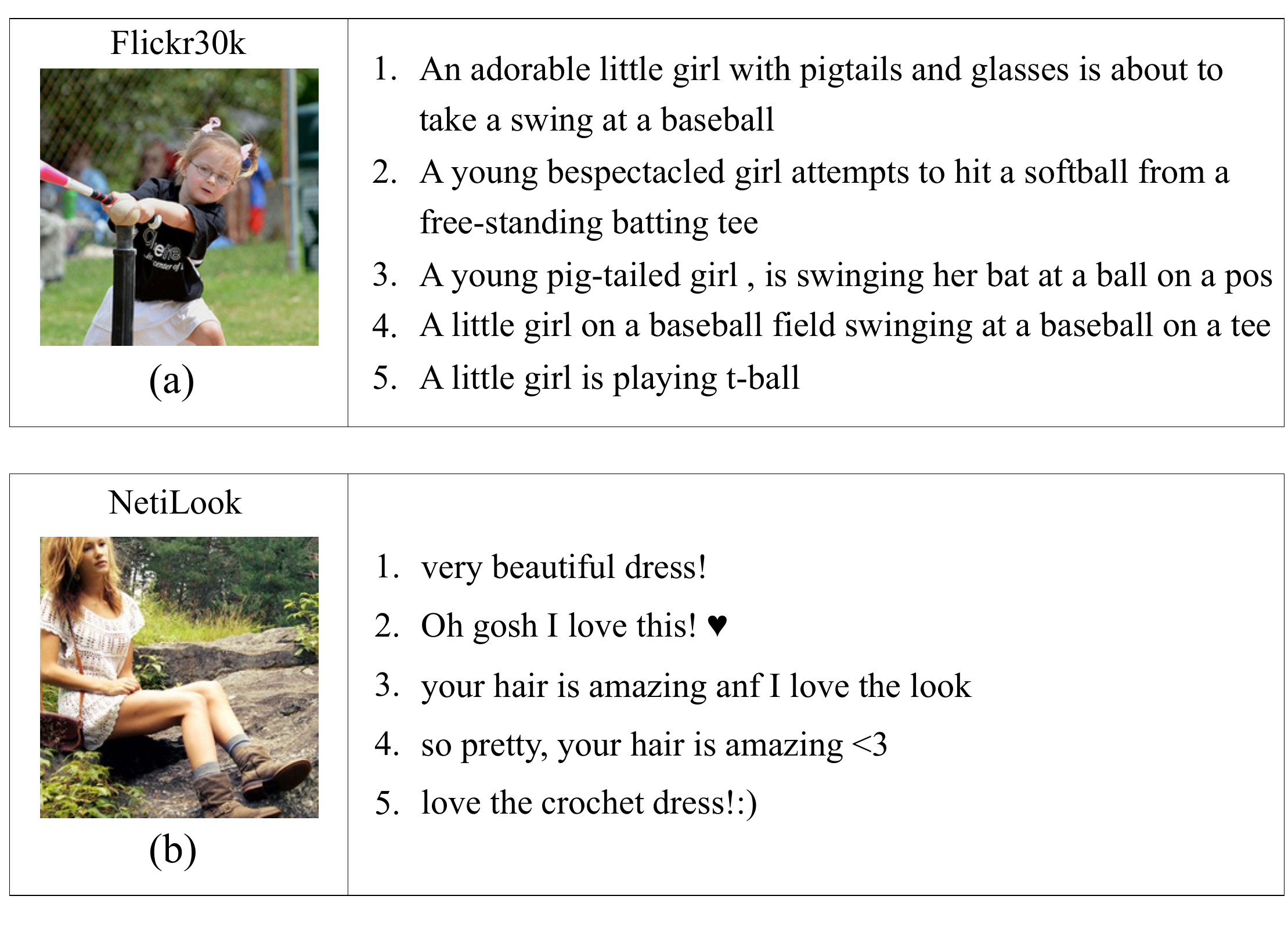}

\caption{Examples from Flickr30k and our NetiLook. (a) Most sentences are describing the shallow appearances (e.g., types, colors) and have similar sentence patterns (e.g., ``A little girl ...''). (b) The sentences involve diverse user intentions with abundant styles. Furthermore, emojis and emoticons inside make it much more intimate with people.
}
\label{fig:dataCompare}
\end{figure}
\section{Dataset --- NetiLook\label{sec:dataset}} 
\begin{figure}
\centering
\includegraphics[width=88mm]{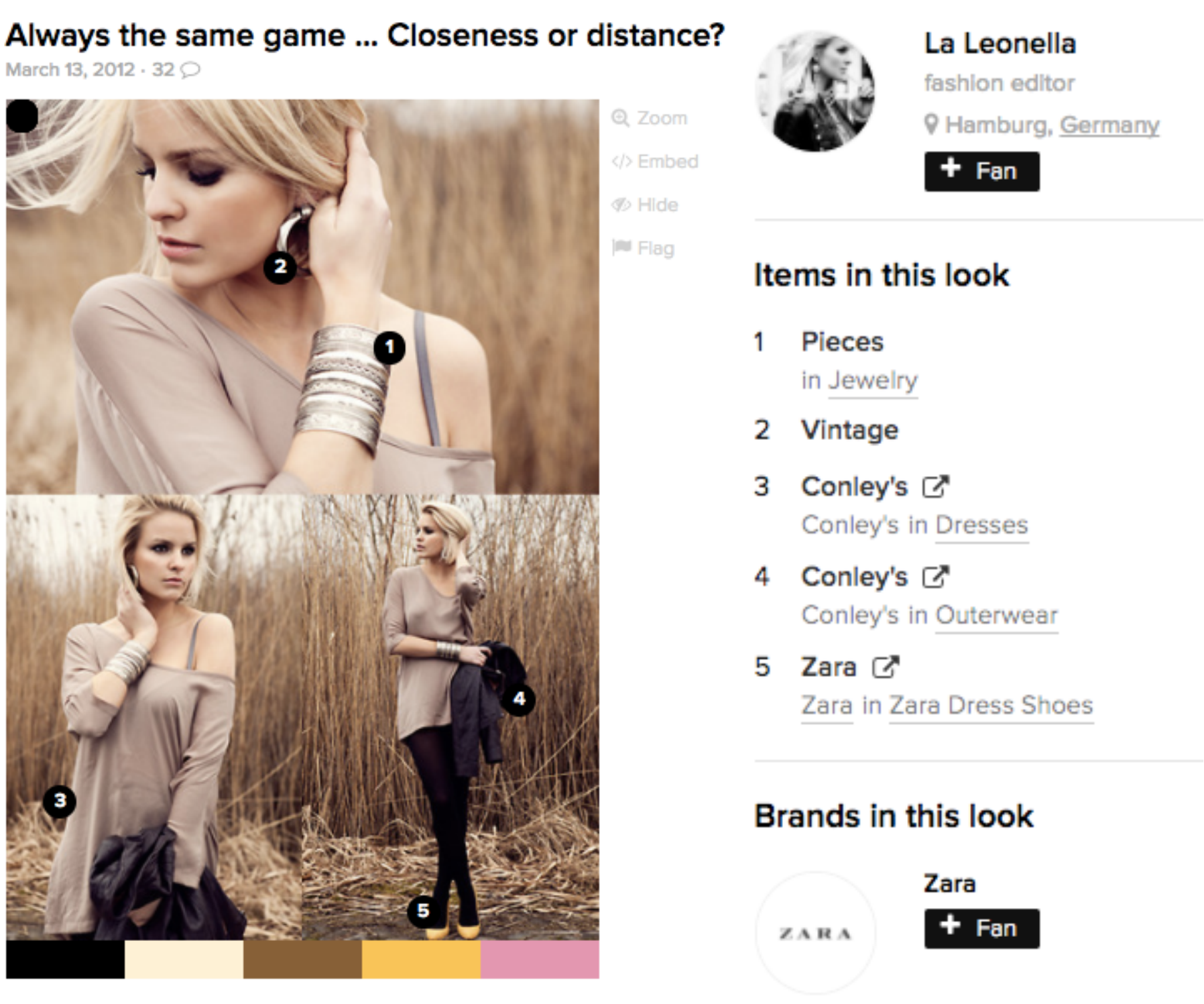}

\caption{An example to show the attributes of a post in Lookbook. The post includes a title named by the author, country of the author, a publishing date, names, brands, and pantone of the clothes.
}
\label{fig:fig5example}
\end{figure}

   \cite{anne2016deep} mentioned that current captioning datasets are relatively small compared with object recognition datasets, such as ImageNet~\cite{deng2010does}. Besides, the descriptions require costly manual annotation. With the growing of social media, such as Facebook and Instagram, people constantly share their life with the world. Consequently, these are all potentially valuable training data for image captioning (or commenting). Among social platforms, there are some specific websites just for clothing style. Lookbook\footnote{lookbook.nu}, an example shown in Figure~\ref{fig:fig5example}, is an online clothing style community where members share their personal style and draw fashion inspiration from each other. Such a rich and engaging social medium is potential to benefit intelligent and human-like commenting applications. Hence, we collected a large corpus of paired user-contributed fashion photos and comments from Lookbook called {\it NetiLook}.

    {\it NetiLook\footnote{https://mashyu.github.io/NSC}}: To the best of our knowledge, this is the first and the largest netizen-style commenting dataset. It contains 355,205 images from 11,034 users and 5 million associated comments collected from Lookbook. As the examples shown in~Figure~\ref{fig:example1}, most of the images are fashion photos in various angles of views, distinct filters and different styles of collage.  As~Figure~\ref{fig:dataCompare} (b) shows, each image is paired with (diverse) user comments. The maximum number of comments is 427 and the average number of comments is 14 per image in our dataset. Note that we observe that there are 7\% of images with no comments and we remove these images in our training stage.  Besides, each post has a title named by an author, a publishing date and the number of hearts given by other user. Moreover, some users add names, brands, pantone of the clothes, and stores where they bought the clothes. Furthermore, we collect the authors' public information. Some of them contain age, gender, country and the number of fans (cf.,~Figure~\ref{fig:fig5example}). We believe all of these are valuable to boost the domain of fashion photo commenting. In this paper, we only use the comments and the photos from our dataset. Other attributes can be used to refine the system in future work. For comparing the results on Flickr30k, we also sampled 28,000 for training, 1,000 for validation and 1,000 for testing. Besides, we also sampled five comments for each image.

\begin{table}[t]
 \tabcolsep=2pt
 \caption{Comparison with other image captioning benchmarks (Flickr30k~\cite{rashtchian2010collecting} and MS COCO~\cite{lin2014microsoft}). Our collected dataset, Netilook, has the most diverse and realistic sentences in the social media (e.g., largest unique words)}
    \label{tab:datacompare}
  \begin{tabular*}{8.5cm}{lcccc}
    \hline
    Dataset & Images& Sentences& 
    \begin{tabular}{@{}c@{}}Average Length\end{tabular}& 
    \begin{tabular}{@{}c@{}}Unique Words\end{tabular}\\
    \hline
  \hline
  Flickr30k& 30K& 150K&13.39 &23,461\\
    \hline
    MS COCO & 200K& 1M&10.46 &54,231\\
  \hline
    NetiLook& 350K& 5M&3.75 & 597,629\\
    \hline
  \end{tabular*}
\end{table}

  Compared to general image captioning datasets such as Flickr30k \cite{rashtchian2010collecting}, the data from social media are quite noisy, full of emojis, emoticons, slang and much shorter (cf.,~Figure~\ref{fig:dataCompare} (b) and~Table~\ref{tab:datacompare}), which makes generating a vivid ``netizen'' style comment much more challenging. Moreover, plenty of photos are in different styles of collage (cf., photos in~Figure~\ref{fig:example1}). Therefore, it makes the image features much more noisy than single view photos. To completely generate comments that entirely reflect the culture in social media, we demonstrate our method in the following section.

\begin{figure*}
\centering
\includegraphics[width=180mm]{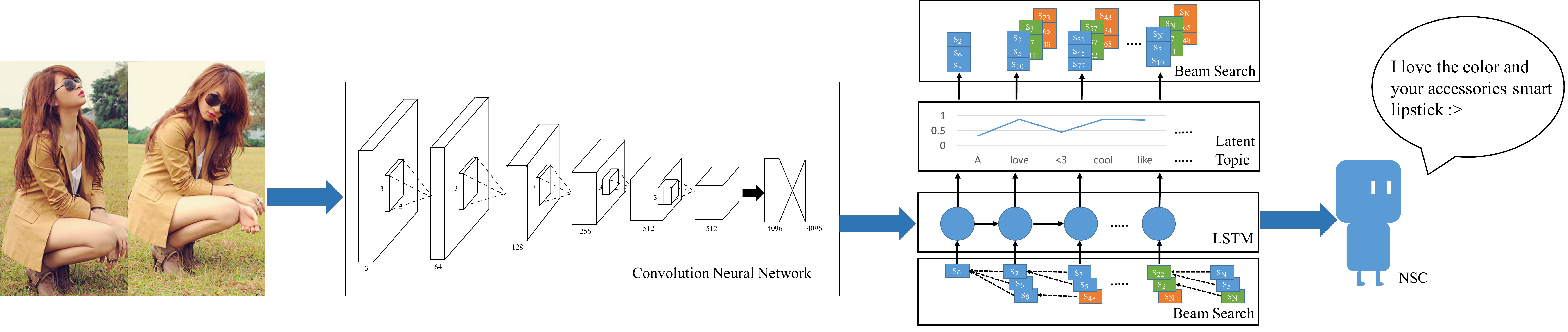}
\caption{An illustration of our proposed framework.
 Our system consists of LSTM~(cf., Section~\ref{sec:imageCaptioning}), topic models~(cf., Section~\ref{sec:styleCaptioning}) and beam search to boost results~(cf., Section~\ref{sec:experimentSetting}). Our proposed approach leverages the outputs of image captioning model based on CNN-RNN frameworks and style-weight from LDA to generate a diverse comment with vivid ``netizen'' style.}
\label{fig:structure}
\end{figure*}

\section{Method -- Netizen Style Commenting}
\label{sec:methods}
In NetiLook, we observed that user comments are much more diverse while comparing them with the sentences in general image captioning datasets. In addition, there are some frequently used sentences along with posts (e.g., ``love this!'', ``nice'') which cause current models inclined to generate similar sentences. The output comments become meaningless and insipid. To immerse the model in vivid netizen style, we fuse style-weight from topic models to image captioning models in order to keep long-range dependencies and take different comments from distinct points of view as topics.

\subsection{Image Captioning}
\label{sec:imageCaptioning}
We follow \cite{vinyals2015show} to extract image features from an image I by a CNN and feed it into the image captioning model at $t=-1$ to inform a LSTM (cf., CNN in~Figure~\ref{fig:structure}). We extract the FC7 (a fully-connected layer) features as high-level meaning of the image from and feed it into the LSTM.
\begin{eqnarray}
\mathbf x_{-1} = \textrm{CNN}( \textrm{I} )\textrm{.}
\end{eqnarray}
We represent a word as a one-hot vector $\mathbf s$ of dimension equal to the size of dictionary. $T$ is the maximum length of output sentences. We represent word embeddings as $W_e$.
\begin{eqnarray}
\mathbf x_t = W_e\mathbf s,~t \in {0...T-1}\textrm{.}
\end{eqnarray}
With the CNN features, we can obtain probabilities of words in each generating step from the image captioning model. Sentences from general image captioning dataset basically depict common content of images. Therefore, conventional image captioning models are able to focus on accuracy. Nevertheless, to strike a balance between accuracy and diversity in current frameworks is arduous. To keep the merit of conventional models, we modify the generating processes of modern models with topic models and make outputs diverse while facing vivid netizen comments.

\subsection{Style Captioning}
\label{sec:styleCaptioning}
To consider vivid netizen style comments, we introduce style-weight $\mathbf{w}_{style}$ element-wised multiplied ($\circ$) with outputs at each step of LSTM to season generated sentences.

\begin{eqnarray}
\label{eq:topiclstm}
\mathbf p_{t+1} = \textrm{Softmax}(\textrm{LSTM}(\mathbf x_t)) \circ \mathbf w_{style},~t \in {0...T-1}\textrm{.}
\end{eqnarray}
Style-weight $\mathbf{w}_{style}$ represents the comment style, which teaches models to be acquainted with style in the corpus while generating captioning. 

However, abstract concepts are hard for people to give a specific definition. To obtain the comment style in NetiLook, we apply LDA to discover latent topics and fuse with current captioning models. 

Suppose, a corpus contains $M$ comments. Comments are composed of a subset of $N$ words. We specify $K$ ($K$ topics ($z_1$, $z_2$, ..., $z_K$)) for LDA. It gives $N$ dimensional topic-word vectors and $K$ dimensional comment-topic vectors.

Topic-word vectors: Each topic {\it z} has a probabilistic vector of $N$ words in dictionary. The vector describes the word distribution of the topic. The topic-word vector $\phi_{z}$ of topic $z$ is
\begin{eqnarray}
\mathbf{\phi_{z}}=\{P(w_1|z),P(w_2|z),...,P(w_N|z)\}\textrm{.}
\end{eqnarray}
where $w_1$, $w_2$, ..., $w_N$ are N words in dictionary.

Comment-topic vectors: Each comment $m$ is also associated with a probabilistic vector of topics, which means the topics probability of the comment. The comment-topic vectors $\theta_{m}$ of comment $m$ is
\begin{eqnarray}
\mathbf{\theta_{m}} = \{P(z_1|m),P(z_2|m),...,P(z_k|m)\}\textrm{.}
\end{eqnarray}
where $z_1$, $z_2$, ..., $z_K$ are different K topics.
 To find the topic distribution in corpus, each comment votes the topic with highest probability by $\arg\max(\theta_m)$. $\mathbf t_m^i$ is the i-th dimension of $\mathbf t_m$. In our finding, the voting gives the most characteristic style in the corpus. The mathematical notation can be represented as follow:
\begin{eqnarray}
Let~\mathbf t_m^i = \{ \begin{array}{l} 
        1~\textrm{if } i = \arg\max(\mathbf{\theta_{m}})\\
        0~\textrm{otherswise}
    \end{array}\textrm{.}
\end{eqnarray}
The topic distribution of the corpus $\mathbf{y}$  now can be computed by normalizing the summation of the number of topics from $ \mathbf t_m$ by the total number of comments. It means the proportion of various points of view of comments in the corpus:
\begin{eqnarray}
\mathbf y = \sum_{m=1}^{M}\mathbf t_m / M\textrm{.}
\end{eqnarray}
With the topic distribution of corpus $\mathbf{y}$ and topic-word vectors $\mathbf{\phi}$, our style-weight $\mathbf{w}_{style}$ is now defined as:
\begin{eqnarray}
\mathbf{w}_{style} = \sum_{k=1}^{K} \mathbf y^k \mathbf{\phi}_k\textrm{.}
\end{eqnarray}
 where $\mathbf y^k$ means the k-th dimension of $\mathbf y$.

As we embed style-weight in~Equation~(\ref{eq:topiclstm}), which could guide the generating process to select words that are much closer to the netizen style learned in the social media (e.g., we observe that one style-wight highlights emoji style), LSTM is capable to generate the sentences with the style in corpus.  (cf., Latent Topic in~Figure~\ref{fig:structure}).  

\section{Diversity Measures}
\label{sec:diversityMeasures}
Since BLEU and METEOR are not for diversity measure, diversity measures are being put importance on sentence generation models. Currently,~\cite{li2015diversity} and~\cite{vijayakumar2016diverse} report the degree of diversity by calculating the number of distinct words in generated responses scaled by the total number of generated tokens. However, this is not enough for diverse comments from the Internet, since comments can be represented not only in natural language but in various sentence patterns, such as emojis, and emoticons. Therefore, to compensate the defects of BLEU and METEOR, we propose three novel measures to judge the diversity of comments generated from captioning models. 

We observed that more diverse sentences are generated, more unique words are used. Thus we devise an intuitive and trivial unique words measure, called DicRate.

{\it DicRate}: The dictionary rate we proposed in this paper is measured through counting number of unique words among generated sentences divided by unique words among ground truth sentences. The number of unique words in ground truth sentences is $N_t$. The number of unique words in generated sentences is $N_g$. The DicRate is computed as follow:
\begin{eqnarray}
\textrm{DicRate}(N_t, N_g) =  N_g / N_t\textrm{.}
\end{eqnarray}
DicRate reflects the abundance of vocabulary of a model, but it is still not incapable to measure sentence diversity.
Inspired by the paper \cite{shao2017generating} for conversation response generation, two novel measures based on entropy are carried out to judge the diversity of comments on fashion photos. Descriptions of the measures are as follows:

{\it WF-KL}: The Kullback-Leibler divergence (KL divergence) of word frequency distribution between ground truth sentences and generated sentences. It  shows how well a model learned the tendency of choosing words in a dataset. The number of unique words in the dataset is $N$. The occurrence times of each word in ground truth sentences are $\mathbf w_t$. The word frequency distribution of ground truth sentences is $\mathbf w_{ft}$. The occurrence of each word in generated sentences are $\mathbf w_g$. The word frequency distribution of generated sentences is $\mathbf w_{fg}$. By referring to the formula of term frequency-inverse document frequency (tf-idf), to avoid division by zero, we add one to $\mathbf w_t$ and $\mathbf w_g$. $\mathbf w^i$ is the i-th dimension of $\mathbf w$.
\begin{eqnarray}
\mathbf w_{ft}^i = (\mathbf w_{t}^i + 1)/ \sum_{i=1}^{N}(\mathbf w_{t}^i+1)\textrm{.}\\
\mathbf w_{fg}^i = (\mathbf w_{g}^i + 1) / \sum_{i=1}^{N}(\mathbf w_{g}^i+1)\textrm{.}
\end{eqnarray}
The WF-KL can be computed as follow:
\begin{eqnarray}
\textrm{WF-KL}(\mathbf w_{ft}, \mathbf w_{fg}) = \sum_{i=1}^{N}\mathbf w_{ft}^i\log(\mathbf w_{ft}^i/\mathbf w_{fg}^i)\textrm{.}
\end{eqnarray}
{\it POS-KL}: The KL divergence of part-of-speech (POS) tagging frequency distribution between ground truth sentences and generated sentences. POS is a classic natural language processing task. One of the applications is identifying which spans of text are products in user search queries~\cite{huang2015bidirectional}. Besides word distribution, POS also demonstrates the interaction between words in a sentence. The number of unique tags in the dataset is $N$. The occurrence times of each tag in ground truth sentences are $\mathbf t_t$. The tag frequency distribution of ground truth sentences is $\mathbf t_{ft}$. The occurrence times of each tag in generated sentences are $\mathbf t_g$. The tag frequency distribution of generated sentences is $\mathbf t_{fg}$. To avoid division by zero, we also add one to $\mathbf t_t$ and $\mathbf t_g$. $\mathbf t^i$ is the i-th dimension of $\mathbf t$.
 \begin{eqnarray}
\mathbf t_{ft}^i = (\mathbf t_{t}^i + 1)/ \sum_{i=1}^{N}(\mathbf t_{t}^i+1)\textrm{.}\\
\mathbf t_{fg}^i = (\mathbf t_{g}^i + 1) / \sum_{i=1}^{N}(\mathbf t_{g}^i+1)\textrm{.}
\end{eqnarray}
The POS-KL can be computed as follow:
\begin{eqnarray}
\textrm{POS-KL}(\mathbf t_{ft}, \mathbf t_{fg}) =  \sum_{i=1}^{N}\mathbf t_{ft}^i\log(\mathbf t_{ft}^i/\mathbf t_{fg}^i)\textrm{.}
\end{eqnarray}

\section{Experiments}
\label{sec:experiments}

\subsection{Experiment Setting}
\label{sec:experimentSetting}
To our best knowledge, this is the first captioning method that focuses on corpus style and sentence diversity. Generally, current methods are devoted to optimizing machine translation scores. Therefore, we only choose two famous captioning methods for comparison rather than other state-of-the-art methods (e.g.,~\cite{anne2016deep},~\cite{wang2016image}). To demonstrate the improvement of diversity, we apply our style-weight to our baselines.

Dataset: Note that we only adopt Flick30k for our experiments to compare with NetiLook because of the characteristic of Flick30k that mainly depicts humans, which is closer to NetiLook. Additionally, images in Flick30k and NetiLook are all collected from social media, which makes images in a similar domain. 

Pre-processing: We argue that the learning process should be autonomous and leverage the freely and hugely available online social media. To avoid noise, we follow \cite{vinyals2015show} to remove the sentences that contain a word frequency that is less than five times in training set. We also filter the sentences that are more than 20 words in the dataset to reduce advertisement and also make sentence more readable \cite{avglen}. Noted that in order to thoroughly convey users' intention and comment style, we do not remove any punctuation in sentences.

Evaluation: BLEU and METEOR are conventional machine translation scores, which base on the matching of answers without considering diversity. The difference between BLEU and METEOR is that METEOR can handle stemming and synonymy matching. In BLEU scores, we report it in 4-grams because it has the highest correlation with humans~\cite{papineni2002bleu}. For BLEU and METEOR, the higher scores mean that sentences are much correct according to the matching with ground truth. In our diversity measures, the higher DicRate shows that the more abundance of vocabulary of a model. Moreover, the lower WF-KL and POS-KL mean that the generated corpus is closer to the ground truth word distribution and sentence patterns.

Baseline: We duplicate two famous captioning methods (NC~\cite{vinyals2015show} and Attention~\cite{xu2015show}) in~Table~\ref{tab:flickr} and~Table~\ref{tab:netilook}. NC is a CNN-RNN framework method that considers global features of images. Attention is an attention-based method, which puts distinct weights on salient features. By comparing NC with Attention in~\cite{xu2015show}, BLEU and METEOR have similar relation like the result reported in~Table~\ref{tab:flickr}. Our proposed method, NSC, fuses style-weight in the decoding stage. Following \cite{vinyals2015show}, we adopt beam search, an approximate inference algorithm, which is widely used in image captioning to boost the results. Because the number of possible sequences grows exponentially with the sentence length, beam search can explore generating process by spreading the most promising node in a limited set. We compare various beam sizes in our experiments and these methods get the best performance at the beam size of 3. Note that the optimal beam size might vary due to the properties of a dataset \cite{kannan2016smart}. In our experiments, for LDA, analysis of the performance sensitivity is made by varying K from 1 to 15. For the first experiment on Flickr 30k, we set the number of topics to be 3 (K = 3); for the experiment on NetiLook, we have K = 5 in $\textrm{NSC}_\textrm{\scriptsize{NC}}$ and K = 3 in $\textrm{NSC}_\textrm{\scriptsize{Attention}}$. We observe that topic models can reflect some semantic ``style'' of comments (e.g. emoji style). Therefore, compared to Flickr 30k, more topic models are selected in NetiLook because user comments are much more diverse in this dataset. Interestingly, the proper number of topic models in $\textrm{NSC}_\textrm{\scriptsize{NC}}$ is higher than $\textrm{NSC}_\textrm{\scriptsize{Attention}}$. We observe that more topic models would not benefit the attention-based approach for the reason that attention-based models are greatly restricted the word selection.

\subsection{Quantitative Analysis -- Dataset}

Traditional captioning dataset such as Flickr30k \cite{rashtchian2010collecting} and MS COCO \cite{lin2014microsoft} only focus on image description and do not emphasize on style and comment-like sentences. Therefore, we address the problem in the paper and contribute the dataset for brand-new problem definition. For comparing models with human and characteristics of datasets, we not only evaluate the generated sentences but also evaluate human comments. Also, as we can see differences from human evaluation between~Table~\ref{tab:flickr} and~Table~\ref{tab:netilook}, the comparison does highlight the distinctions between Netilook and Flicr30k. For a comment given by a human or machine, it is difficult to be evaluated on conventional measures such as BLEU in NetiLook (e.g.~0.108 in~Table~\ref{tab:flickr} vs. 0.008 in~Table~\ref{tab:netilook} in BLEU-4). Thus, we propose our measures DicRate, WF-KL and POS-KL to evaluate comments.

\begin{table}[t]
  \label{tab:commands}
  \tabcolsep=2pt
  \caption{Performance on Flickr30k testing splits.}
  \label{tab:flickr}
  \begin{tabular*}{8.5cm}{p{1.3cm}<{\centering} p{1.3cm}<{\centering} p{1.3cm}<{\centering} p{1.3cm}<{\centering} p{1.3cm}<{\centering} p{1.3cm}<{\centering}}
    \hline
    Method&       BLUE-4&   METEOR& WF-KL& POS-KL& DicRate \\
    \hline
    \hline
    Human&      0.108& 0.235& 1.090& 0.013& 0.664\\
    \hline
    NC&            0.094& 0.147& 1.215& 0.083& 0.216 \\
    Attention&     \textbf{0.121}& \textbf{0.148}& 1.203& 0.302& 0.053\\
    \hline
    \hline
    $\textrm{NSC}_\textrm{\scriptsize{NC}}$&  0.089& 0.146& 1.217& \textbf{0.075}& \textbf{0.228}\\
    $\textrm{NSC}_\textrm{\scriptsize{Attention}}$&  0.119& \textbf{0.148}& \textbf{1.202}& 0.319& 0.055\\
    \hline
  \end{tabular*}
\end{table}

\begin{table}[t]
  \label{tab:commands}
  \tabcolsep=2pt
   \caption{Performance on NetiLook testing splits.}
  \label{tab:netilook}
  \begin{tabular*}{8.5cm}{p{1.3cm}<{\centering} p{1.3cm}<{\centering} p{1.3cm}<{\centering} p{1.3cm}<{\centering} p{1.3cm}<{\centering} p{1.3cm}<{\centering}}

    \hline
    Method&       BLEU-4&   METEOR& WF-KL& POS-KL& DicRate \\
    \hline
    \hline
    Human&        0.008& 0.172& 0.551& 0.004& 0.381\\
    \hline
    NC&           0.013& 0.151& 0.665& 1.126& 0.036\\
    Attention&     0.020& 0.133& \textbf{0.639}& 1.629& 0.011\\
    \hline
    \hline
    $\textrm{NSC}_\textrm{\scriptsize{NC}}$&   0.013& \textbf{0.172}& 0.695& \textbf{0.376}& \textbf{0.072}\\
    $\textrm{NSC}_\textrm{\scriptsize{Attention}}$&  \textbf{0.030}& 0.139& 0.659& 1.892& 0.012\\
    \hline
  \end{tabular*}
 \end{table}

\begin{figure*}[h!]
\centering
\includegraphics[width = 179mm]{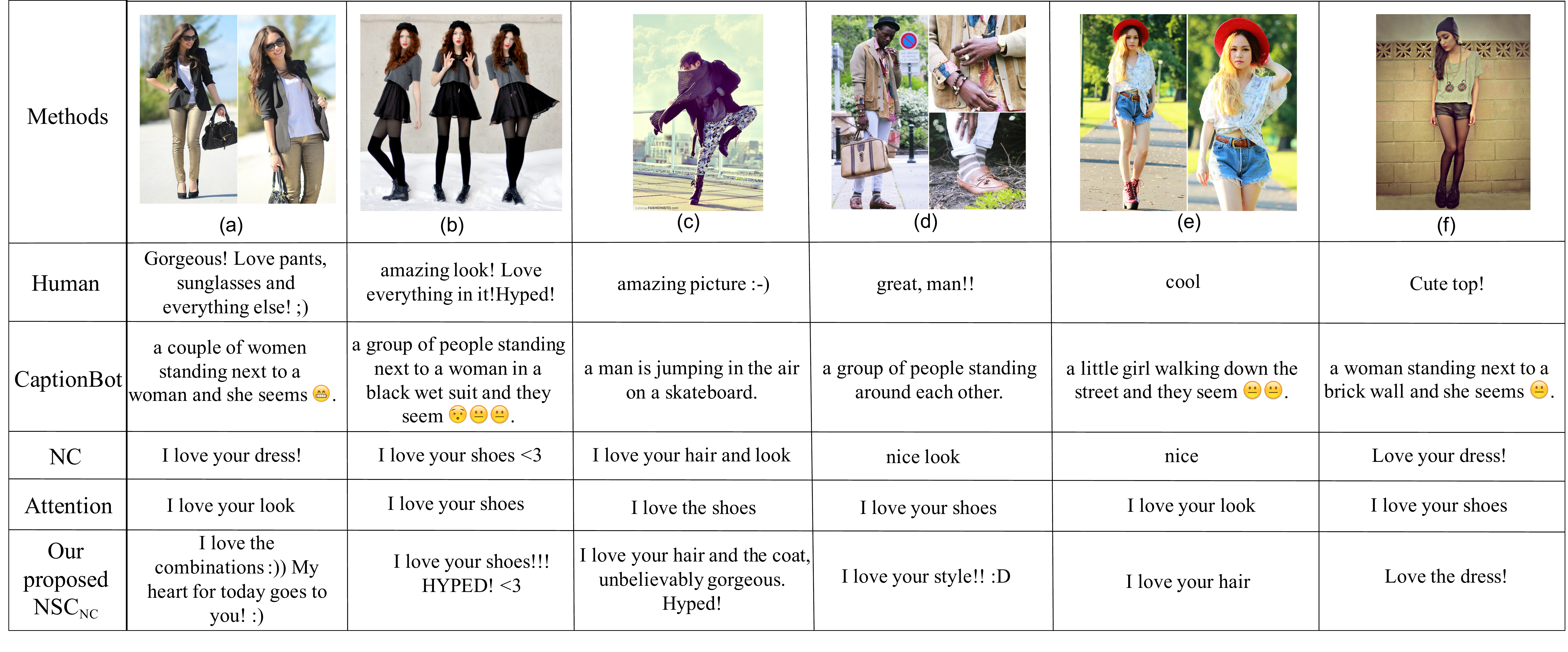}
\caption{Examples of comments generated by different methods.
    The examples show that our proposed approach $\textrm{NSC}_\textrm{\scriptsize{NC}}$ can help generate more diverse and vivid comments.}
\label{fig:examples}

\end{figure*}

In the scenario of online social media, punctuation, slang, emoticons and emojis are important for conveying emotion in a sentence. Thus, Netilook has much more diversity and unique words than other datasets (0.664 in~Table~\ref{tab:flickr} vs. 0.381 in~Table~\ref{tab:netilook} in DicRate). NetiLook specializes in describing clothing style as examples shown in~Figure~\ref{fig:examples}. Still, there are some common words and general patterns to describe and comment on the clothing style in comparison with Flickr30k, which mixes all types of images in the dataset. Thus NetiLook has lower score on WF-KL and POS-KL (e.g.~1.090 in~Table~\ref{tab:flickr} vs. 0.551 in~Table~\ref{tab:netilook} in WF-KL).

For such a diverse and characteristic dataset, machines are required considering overall corpus distribution and mimic comment style in order to get high performance in our evaluations. Nonetheless, for learning human commenting style, it is still challenging for general captioning models to generate diverse words while there are some general comments can achieve universally low loss (e.g., ``nice'', ``I love this!''). However, our style-weight brings human style in machine generated sentences.

\subsection{Quantitative Analysis -- Model Evaluation}
 Table~\ref{tab:flickr} summarizes performances for the Flickr30k dataset. Attention models put weights on salient features in images, thus the models easily describe objects inside pictures and reach a better BLEU and METEOR (e.g.~0.094 vs. 0.121 in BLEU-4). However, attention-based models are greatly restricted the word selection while decoding stage. In our experiments, POS-KL and DicRate are much worse (e.g.~0.053 vs. 0.216 in DicRate) comparing Attention with NC. With our style-weight, the model $\textrm{NSC}_\textrm{\scriptsize{NC}}$ expands the word diversity without sacrifice much on BLEU and METEOR. Style-weight encourages models choosing the words that are closer to the original distribution rather than the words that can generally get the lowest loss during the training phase. As we show in~Table~\ref{tab:flickr}, DicRate and POS-KL are improved comparing $\textrm{NSC}_\textrm{\scriptsize{NC}}$ with NC (e.g.~0.216 vs. 0.228 in DicRate). The impact of style-weight is also shown in Attention model. However, we observed that embedding style-weight does not improve much in Flickr30k dataset in terms of diversity, because the sentences are objectively depicting humans performing various activities in Flickr30k. 
 
 In NetiLook, the experiment in~Table~\ref{tab:netilook} shows that our method can greatly improve the diversity. Comparing NC with Attention in~Table~\ref{tab:netilook}, NC performs better than Attention (e.g.~0.036 vs. 0.011 in DicRate) except for BLEU-4 and WF-KL (e.g.~0.665 vs. 0.639 in WF-KL) because the selection of words is affected by salient features in an image, which makes the model miss the intention of the corpus while the whole dataset has similar objects. However, with style-weight, our $\textrm{NSC}_\textrm{\scriptsize{NC}}$ outperforms other baselines in POS-KL and DicRate (e.g.~0.376 of $\textrm{NSC}_\textrm{\scriptsize{NC}}$ in POS-KL). This proves that style-weight can guide the generating process to the comment that is much closer to the users' behaviour in the social media, making machine mimic online netizen comment style.

\subsection{Image Commenting Results}

 We show some real examples of fashion commenting results on NetiLook with various methods. Though there are emojis generated from Microsoft CaptionBot, the comments are still lacking engagement and can not afford to process photos in collage. While training general captioning models (e.g., NC and Attention) on NetiLook, the comments are much shorter than Human's and fixed in some patterns, which lacks diversity.

{\it Similar intention like human}: With the style-weight, $\textrm{NSC}_\textrm{\scriptsize{NC}}$ can generate the comment that is much closer to users' intention (cf.,~Figure~\ref{fig:examples} (a)).

{\it More vivid comments}: While conveying the same intention, $\textrm{NSC}_\textrm{\scriptsize{NC}}$ is able to use emoticons, punctuations and capitalizations to generate more netizen-style comments than other captioning models (cf.,~Figure~\ref{fig:examples} (b)).

{\it Another point of view}: By considering the topic distribution of data, $\textrm{NSC}_\textrm{\scriptsize{NC}}$ generates comments that are different from general captioning models' and much closer to human beings (cf.,~Figure~\ref{fig:examples} (c) - (e)).

{\it Wrong objects}: However, there are still some drawbacks in our $\textrm{NSC}_\textrm{\scriptsize{NC}}$, such as describing wrong objects in the images. Because the $\textrm{NSC}_\textrm{\scriptsize{NC}}$ is still based on image captioning models, it will also generate wrong comments as other captioning models due to the similarity of images (cf.,~Figure~\ref{fig:examples} (f)). It can be improved by jointly training the topic model with attention-based models.

\subsection{User Study}
Motivated by the paper ~\cite{wang2016image} which conducts a human evaluation of image captioning by presenting images to three workers, we conducted a user study from 23 users to demonstrate the effect of diverse comments. The users are about 25 year-old and familiar with netizen style community and social media. The sex ratio in our user study is 2.83 males/female. They are asked to rank comments for 35 fashion photos. Each photo has 4 comments --- from one randomly picked human comments, NC, Attention and our $\textrm{NSC}_\textrm{\scriptsize{NC}}$. Therefore, each of the users has to appraise 140 comments generated from different methods. Furthermore, we collect user feedback to understand user judgements on comments generated by different methods.

As~Table~\ref{tab:userStudy} shows, 36.8\% out of 805 votes ($35\times 23$) rank the sentences generated from $\textrm{NSC}_\textrm{\scriptsize{NC}}$ at the first place which outperforms NC and Attention. It means that our NSC defeats human comments in some images. 
Furthermore, The difference between humans and $\textrm{NSC}_\textrm{\scriptsize{NC}}$ in rank 1 is only 9.3\%. In top two ranks, the performance of $\textrm{NSC}_\textrm{\scriptsize{NC}}$ reaches 76.6\%. This also demonstrates that our $\textrm{NSC}_\textrm{\scriptsize{NC}}$ can generate sentences with human-like quality. In our user study, people generally regard our $\textrm{NSC}_\textrm{\scriptsize{NC}}$ sentences as human comments. According to our user study, the main concern of people's ranking is emoticons. Emoticons is an important component in the sentences to connect human emotions and also make sentences more vivid. For instance, Figure~\ref{fig:examples} (d) in the user study, the voting of $\textrm{NSC}_\textrm{\scriptsize{NC}}$ outperforms Human at the first rank (39.1\% vs. 34.8\%). Relevance between comments and images takes the second concern of people's ranking. Objects mentioned in sentences should not be trivial or mismatch in the photos. For example, Figure~\ref{fig:examples} (c), $\textrm{NSC}_\textrm{\scriptsize{NC}}$ captures the outfit (coat) and floating hair resulting in the same voting as human in rank 1 (39.1\%) in the user study. To sum up, our style-weight makes captioning model mimic human style and generates human-like comments which most people agree with in our user study.

\begin{table}[t]
  \label{tab:commands}
  \centering
   \caption{Result of user study. $\textrm{NSC}_\textrm{\scriptsize{NC}}$'s comments are more likely to be regarded as human than other methods.}
    \label{tab:userStudy}
  \begin{tabular*}{8.5cm}{p{1.3cm}<{\centering} p{1.3cm}<{\centering} p{1.3cm}<{\centering} p{1.3cm}<{\centering} p{1.3cm}<{\centering} p{1.3cm}<{\centering}}
    \hline
    Ranking  & Human& NC& Attention &$\textrm{NSC}_\textrm{\scriptsize{NC}}$\\
    \hline
    \hline
  Rank 1& 46.1\%& 10.8\%& 6.3\%& 36.8\%\\
    \hline
    Rank 2& 24.5\%& 21.4\%&14.4\%& 39.8\%\\
    \hline
    Rank 3& 18.1\%&31.9\%&34.3\%& 15.7\%\\
    \hline
    Rank 4& 11.3\%& 35.9\%& 45.0\%& 7.8\%\\
    \hline
  \end{tabular*}
\end{table}

\section{Conclusions}
We present style-weight that greatly influences on current captioning models to immerse into human online society. Also, we contribute our dataset NetiLook, which is a brand new large-scale clothing dataset, to achieve netizen style commenting with style-weight. An image captioning model automatically generates characteristic comments for user-contributed fashion photos. NSC leverages the advantage of style-weight which can keep long-range dependencies to achieve vivid ``netizen'' style comments. Experiments on Flickr30k and NetiLook datasets, we demonstrate our proposed approaches benefit fashion photo commenting and improve image captioning task both in accuracy (quantified by conventional measures of image captioning) and diversity (quantified by our proposed measures). A user study is carried showing that our proposed idea can generate sentences with human-like quality. It is worth noting that our proposed approach can be applied on other fields (e.g., conversation response generation or question-answering system) to help generate sentences with various styles by the idea of style-weight. Moreover, NetiLook contains abundant attributes, researchers are able to use those attributes to build a more comprehensive system. For example, comments from different genders in future work. We believe that the integration of image captioning models, style-weight and the dataset proposed in this paper will have a great impact on related research domains.

\section{Acknowledgement}
This work was supported in part by Microsoft Research Asia and the Ministry of Science and Technology, Taiwan, under Grant MOST 105-2218-E-002-032. We also benefit from the grants from NVIDIA and the NVIDIA DGX-1 AI Supercomputer and the discussions with Dr. Ruihua Song, Microsoft Research Asia.

\bibliographystyle{ACM-Reference-Format}
\bibliography{sample-bibliography} 

\end{document}